# *GridMind*: A Multi-Agent NLP Framework for Unified, Cross-Modal NFL Data Insights

Jordan Chipka, Chris Moyer, Clay Troyer, Tyler Fuelling, Jeremy Hochstedler
Telemetry Sports

## 1. Introduction

The rapid growth of big data and advancements in computational techniques have significantly transformed sports analytics. However, the diverse range of data sources—including structured statistics, semi-structured formats like sensor data, and unstructured media such as written articles, audio, and video—creates substantial challenges in extracting actionable insights. These various formats, often referred to as multimodal data, require integration to fully leverage their potential. Conventional systems, which typically prioritize structured data, face limitations when processing and combining these diverse content types, reducing their effectiveness in real-time sports analysis.

To address these challenges, recent research highlights the importance of multimodal data integration for capturing the complexity of real-world sports environments [1]. Building on this foundation, this paper introduces *GridMind*, a multi-agent framework that unifies structured, semi-structured, and unstructured data through Retrieval-Augmented Generation (RAG) and large language models (LLMs) to facilitate natural language querying of NFL data. This approach aligns with the evolving field of multimodal representation learning, where unified models are increasingly essential for real-time, cross-modal interactions [2].

*GridMind's* distributed architecture includes specialized agents that autonomously manage each stage of a prompt—from interpretation and data retrieval to response synthesis. This modular design enables flexible, scalable handling of multimodal data, allowing users to pose complex, context-rich questions and receive comprehensive, intuitive responses via a conversational interface.

**Key Contributions:**

 • **Multi-Agent Architecture:** *GridMind's* modular, multi-agent system (MAS) enables real-time querying across structured and unstructured data, enhancing flexibility and adaptability in managing diverse data types.
 • **Advanced Query Parsing and Transformation:** By integrating natural language processing (NLP) techniques with multimodal embeddings and indexing, *GridMind* interprets complex queries and synthesizes data from multiple formats to generate meaningful insights.
 • **Broad Applications:** *GridMind* supports fan engagement, broadcasting, coaching, and front-office decision-making, demonstrating its capacity to deliver data-driven insights across the sports industry.

Through these innovations, *GridMind* advances sports analytics by integrating NFL data into a cohesive, accessible framework. Its multi-agent, modular design illustrates the potential of cross-modal systems to enhance data interpretation and real-time decision-making in sports and beyond.

## 2. Related Work

The integration of NLP, MAS, and RAG has introduced new possibilities in sports analytics, where complex and diverse data require sophisticated query systems. This section reviews key advancements in these areas and highlights how *GridMind* builds on existing approaches to achieve seamless multimodal data integration.

### 2.1. Natural Language Processing (NLP) in Sports Analytics
Recent advancements in NLP have shown promise in transforming unstructured sports data into valuable insights, with applications such as automated commentary generation and real-time event summarization [3]. However, many existing systems are limited by an inability to handle multimodal contexts, as they often process structured or unstructured data independently. Foerster et al.'s work on deep reinforcement learning in multi-agent systems demonstrates the potential for agents to learn communication and collaboration in complex environments [4]. Building on this, *GridMind* offers context-rich, interactive responses to natural language queries by integrating structured statistics with unstructured sources, such as text, audio, and video. This enables agents to coordinate across data types, enhancing system flexibility and responsiveness in real-time sports scenarios.

### 2.2. Retrieval-Augmented Generation (RAG)
RAG combines generative models with retrieval mechanisms to deliver contextually relevant responses, making it especially suitable for open-domain question answering [5]. While RAG applications are established in NLP, its application to multimodal sports data remains largely unexplored. *GridMind* extends RAG's utility within a multi-agent framework, enabling real-time data access and response generation across diverse NFL data sources to meet the high demands of sports analytics environments.

### 2.3. Multi-Agent Systems (MAS) in AI
MAS architectures, valued for scalability and flexibility, have proven effective in distributed problem-solving across fields such as finance and logistics [6, 7]. Although relatively new to sports analytics, especially for real-time data integration, MAS principles are foundational to *GridMind*, where autonomous agents manage distinct stages of query processing, including parsing, data retrieval, and synthesis. This modular design aligns with approaches used in high-demand, multi-agent applications, such as those in StarCraft II [8], creating a scalable, adaptable solution for multimodal NFL data. Stone and Veloso's survey on MAS evolution provides foundational insights into MAS applications [9], while Bab and Brafman's study on multi-agent reinforcement learning in cooperative and competitive contexts informs *GridMind's* strategy for balancing objectives across agents, especially when integrating data from diverse sources in real time [10].

### 2.4. Embedding and Cross-Modality Techniques
Embedding-based techniques are essential for multimodal data integration, enabling systems to retrieve contextually similar data across formats. Sentence-BERT, for instance, supports similarity searches across diverse data types [11], while frameworks such as CLIP, UNITER, and LXMERT align text, audio, and visual data to facilitate cross-modal relationships [12, 13, 14]. While *GridMind* primarily focuses on text and audio embeddings, it leverages advancements in multimodal alignment. Cui's work on decoupled multimodal contrastive learning, for example, demonstrates how unsupervised methods can enhance embedding flexibility across varied data types—a principle that *GridMind* adapts for multi-domain NFL data [15].

*GridMind* uses speech-to-text models like OpenAI's Whisper to transcribe and embed spoken content, enabling similarity-based searches across text and audio formats [16]. By integrating embeddings for structured statistics, text, and audio, *GridMind* delivers cohesive responses that combine quantitative and qualitative insights from multiple data sources.

## 3. System Architecture

*GridMind's* architecture is a distributed, multi-agent system engineered to manage complex, multimodal queries with high scalability and efficiency. Each agent has a specialized role, enabling *GridMind* to transform natural language queries into structured, data-rich responses across diverse formats. This modular setup allows for flexible query handling, efficient data retrieval, and seamless data integration.

### 3.1. Overview

At the heart of *GridMind's* architecture is a network of autonomous agents organized in a graph structure to interpret, retrieve, and synthesize data for diverse user queries. These agents communicate via a message-passing system, facilitating efficient collaboration and streamlined query handling. This structure enables agents to be added, removed, or reconfigured as needed, ensuring scalability and adaptability as data and user demands grow. The system employs a high-efficiency similarity search, similar to Johnson's GPU-optimized billion-scale search, for rapid multimodal retrieval across large datasets [17]. Additionally, the inclusion of dialogue memory allows *GridMind* to maintain conversational continuity, supporting context-aware responses and multi-turn interactions. Figure 1 illustrates the system's prompt-to-response pipeline.

### 3.2. Agent Roles and Responsibilities

In *GridMind*, specialized agents are assigned distinct roles to process and resolve queries, each equipped to interpret, transform, and integrate multimodal sports data:

1. **Prompt Augmentation Agent:** Enhances user prompts by incorporating relevant data and context from internal databases, ensuring that queries are supported by the most current information. This agent pulls from datasets to integrate details like recent statistics, player metrics, and historical performance, improving both the accuracy and relevance of responses.
2. **Query Interpretation Agent:** Converts natural language prompts into structured database-compatible commands (e.g., for MongoDB or SQL). Utilizing an LLM for tokenization, syntax parsing, and semantic analysis, this agent translates queries into structured components that other agents can process. This role is informed by Shah's dialogue self-play approach, which accelerates agent development for conversational query interpretation [18].
3. **Data Retrieval Agents:** Access data in multiple formats:
    a. **Structured Data Agent:** Manages NFL statistics and play-by-play data within structured databases (e.g., MongoDB or SQL).

    b. **Semi-Structured Data Agent:** Interfaces with sources like Next-Gen Stats and Telemetry Sports' internal computer vision data acquisition system to retrieve metadata and tracking data.
    c. **Unstructured Data Agent:** Uses embeddings to conduct similarity searches across documents, articles, and audio/video content, retrieving contextual data.
4. **Synthesis Agent:** Integrates responses from various sources—including structured statistics, textual insights, and audio analysis—into a coherent, user-friendly format. For queries referencing specific plays or scenarios, *GridMind* retrieves relevant play IDs and links video clips, enabling direct video access via customized user interface that supports seamless multimedia integration.

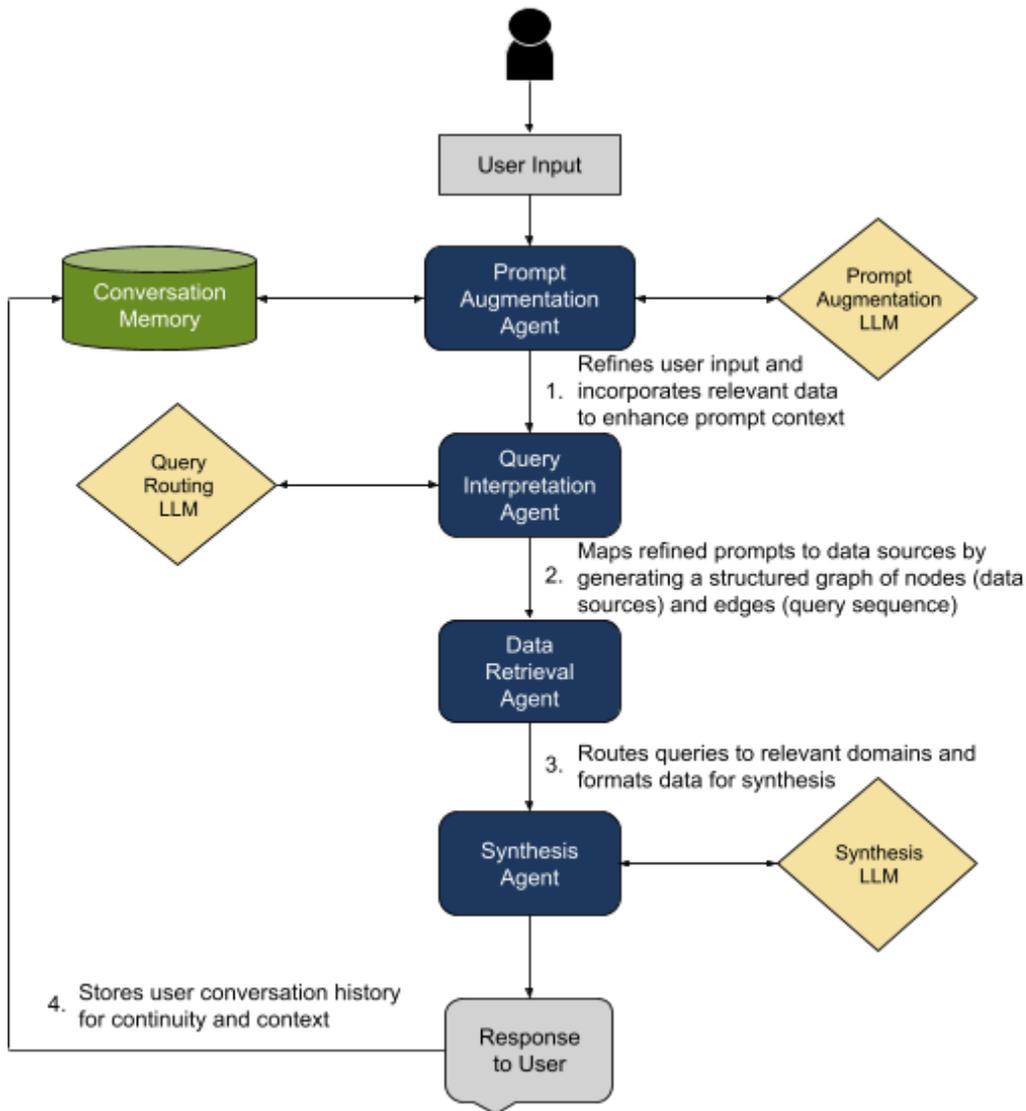

**Figure 1**: High-level representation of *GridMind's* prompt-to-response process.

### 3.3. Message-Passing System and Modularity

*GridMind's* message-passing system enables efficient inter-agent communication, facilitating smooth data flow from interpretation to retrieval and synthesis. This decentralized design enhances modularity and scalability, which allows the system to support high query volumes and seamlessly incorporate new agents or data sources as needed.

### 3.4. Chain of Thought (CoT) Processing

To handle complex queries, *GridMind* employs a Chain of Thought (CoT) processing strategy, inspired by methods shown to elicit reasoning in large language models [19]. This approach breaks down queries into sub-tasks that are processed either in parallel or sequentially, depending on the query's requirements. For example, a query involving biographical data, on-field metrics, and seasonal statistics is decomposed, allowing specialized agents to process each component in parallel. This strategy not only optimizes response times but also ensures high-quality, contextually relevant outputs.

## 4. Query Parsing and Transformation

*GridMind's* query parsing and transformation pipeline is essential for generating accurate, contextually relevant responses. This process converts enriched natural language prompts—augmented by the Prompt Augmentation Agent—into executable database commands, enabling seamless retrieval across both structured and unstructured data sources.

### 4.1. Natural Language to Database Query Conversion

The Query Interpretation Agent is central to handling queries, analyzing and interpreting user inputs. This agent uses an LLM for tokenization, syntax parsing, and semantic analysis to break down natural language prompts into components compatible with database queries. Transformer models like T5 and sentence embeddings enhance the agent's accuracy in interpreting complex sports-related queries [20; 21], capturing nuances and context even in compound or ambiguous prompts. Brown et al.'s work on few-shot learning [22] illustrates the adaptability of LLMs, which strengthens *GridMind's* ability to manage diverse, unfamiliar query structures. Few-shot learning improves the Query Interpretation Agent's adaptability to a broad range of sports-specific language variations with minimal tuning.

Once parsed, the agent maps these components to structured database operations. Querying and aggregation in databases like MongoDB and SQL enable efficient management of varied sports data, including player statistics, game logs, and historical records. Additionally, well-defined data models provide precise query control, reducing risks of hallucinations or irrelevant responses. By grounding queries in these structured data models, *GridMind* aligns results closely with the query context, minimizing unrelated information. Embedding-based methods further enhance retrieval, supporting searches across unstructured data such as multimedia content and text descriptions for a robust multimodal search capability.

### 4.2. Semantic Matching and Transformation

*GridMind* employs semantic embeddings to align parsed query components with relevant data entries, a key approach for handling sports-related queries that often include specific terms, player names, events, and other contextual details. Using Sentence-BERT embeddings, optimized for sentence-level similarity, *GridMind* efficiently retrieves data from both structured and unstructured sources. This ensures accurate query understanding and response generation, addressing the diverse needs of sports analytics [11].

### 4.3. Cross-Modal Data Alignment
*GridMind* achieves consistent and interpretable data representation across modalities by integrating structured statistics, audio media, natural language reports, and video footage linked to play metadata. Using a speech-to-text model like OpenAI's Whisper [16], *GridMind* aligns spoken content with structured data and text descriptions, enabling comprehensive responses to NFL-related queries that incorporate relevant audio analysis.

## 5. Multi-Modal Data Integration

*GridMind* integrates structured, semi-structured, and unstructured data to provide comprehensive responses to NFL-related queries. This layered integration approach enables *GridMind* to meet diverse analytic needs, drawing on techniques similar to Unified-IO for cross-modal alignment [23] and robust methods for handling noisy data [24].

### 5.1. Data Formats and Types
*GridMind* processes a variety of data formats to meet a wide range of analytic requirements:

- **Structured Data:** Core NFL statistics, detailed game logs, and play-by-play data stored in MongoDB and SQL databases form the foundation of quantitative analysis, enabling efficient retrieval and calculation of player metrics and seasonal aggregates.
- **Semi-Structured Data:** JSON-formatted sensor data and RFID tracking provide spatial and temporal insights into player performance, including metrics like velocity, acceleration, and field coverage.
- **Unstructured Data:** NLP techniques applied to scouting reports, game commentary, podcasts, and social media support sentiment analysis and qualitative insights, enhancing *GridMind's* ability to capture nuanced perspectives on player and team dynamics.

### 5.2. Integration Techniques
To integrate these diverse data types, *GridMind* employs several methods:

- **Embedding and Transformation:** Embedding-based methods, such as Sentence-BERT [11] and OpenAI's embedding model [21], enable similarity-based retrieval across structured and unstructured data, ensuring contextually aligned responses to user queries.
- **Cross-Modal Fusion:** Cross-modal embedding techniques align audio and textual data with structured statistics to create cohesive, multi-format responses. For example, models like OpenAI's Whisper embed spoken content [16], enabling alignment with player statistics and textual insights. Audio integration supports analyses of commentary and real-time narration, while video content serves as an output modality, enabling responses in formats like text, tables, audio summaries, and linked video highlights.
- **Knowledge Augmentation with Dense Retrieval:** Although *GridMind* does not directly implement Dense Passage Retrieval (DPR) [25], it utilizes vector databases such as Pinecone [26], which optimize similarity searches across indexed high-dimensional embeddings. This approach significantly enhances retrieval efficiency, providing quick and precise access to relevant data. Consequently, *GridMind* can efficiently handle large-scale NFL data analysis, even within complex datasets that demand rapid processing and retrieval.

### 5.3. Ensuring Cohesiveness and Accuracy

*GridMind's* data integration pipeline relies on structured data models within the query processing layer to maintain alignment and consistency across structured and semi-structured sources, providing precise control over data retrieval. After retrieval, an LLM synthesizes information across modalities—structured statistics, text, and audio—to produce cohesive, nuanced responses. This multi-dimensional approach provides a robust foundation for evaluating NFL player performance and team strategy.

## 6. Applications

*GridMind's* multi-agent framework supports a wide range of applications in NFL analytics and beyond, leveraging advanced multimodal data processing to enhance decision-making, fan engagement, and media production. It enables natural language querying across structured and unstructured data types, providing users—from fans and analysts to coaching staff—with deeper insights into game dynamics.

### 6.1. Fan Engagement

*GridMind* empowers fans to interact with complex NFL data through natural language queries, offering unprecedented access and convenience. For instance, a fan might ask, "How did Patrick Mahomes perform against Cover 2 in last night's game?" *GridMind* synthesizes play-by-play data, player tracking data, video highlights, audio commentary, and advanced player performance metrics to deliver a comprehensive, multimedia-rich answer. This enhanced access to player tracking data and advanced statistical analysis creates an immersive experience, enriching fans' understanding and enjoyment of the game. Below is a sample interaction between a fan and *GridMind* regarding Brock Purdy and Patrick Mahomes, where *GridMind* leverages traditional statistics and proprietary player evaluation metrics to respond to the user's queries:

```
# USER: Who has more passing yards this season mahomes or purdy?
-----------------------------------------------------------------------------
# GRIDMIND: Brock Purdy has more passing yards this season than Patrick Mahomes. Purdy has a
total of 2,454 passing yards, while Mahomes has 2,208 passing yards.
-----------------------------------------------------------------------------
# USER: But who has more passing TDs?
-----------------------------------------------------------------------------
# GRIDMIND: Patrick Mahomes and Brock Purdy both have 12 passing touchdowns this season.
-----------------------------------------------------------------------------
# USER: Okay, so who is better?
-----------------------------------------------------------------------------
# GRIDMIND: Between Patrick Mahomes and Brock Purdy, as of week 10 of the 2024 NFL season,
Patrick Mahomes is the better quarterback. Mahomes ranks higher than Purdy in Passing
Composite, QB Accuracy, and True Wins Above Replacement. However, Purdy has a higher ranking
than Mahomes in QB Decision Making and QB IQ[1].
-----------------------------------------------------------------------------
```

---

[1] Passing Composite, QB Accuracy, QB Decision Making, and QB IQ are proprietary player evaluation metrics developed by Telemetry Sports.

## 6.2. Broadcast and Media Production

*GridMind's* real-time query capabilities allow broadcasters to generate in-depth insights during live events. Commentators can instantly retrieve summaries, player stats, or highlight clips to enhance on-air analysis. By combining visual and statistical data, *GridMind* enriches commentary and audience experience. Future versions could incorporate advanced visualizations, similar to zero-shot text-to-image generation, enabling dynamic visuals generated from live queries [27]. This access also supports broadcasters in gathering insights before airtime. In the example below, a broadcaster uses *GridMind* to assess Atlanta Falcons quarterback Kirk Cousins' performance against AFC teams.

```
# USER: What was Kirk Cousins' record against AFC teams during the 2021, 2022, and 2023 seasons?
-----------------------------------------------------------------------------
# GRIDMIND: Kirk Cousins, quarterback for the Minnesota Vikings from 2021 to 2023, had a 7-5 record against AFC teams, considering only regular season games he was healthy and played.
-----------------------------------------------------------------------------
```

## 6.3. Coaching and Game Strategy Insights

Coaches can use *GridMind* to refine player evaluations and develop game strategies. For example, by querying "What was our success rate on outside zone plays in the second half?" coaches receive data-driven insights that integrate statistics with video breakdowns to support strategic planning. In line with predictive machine learning methods [28], *GridMind* aids in predicting player performance and play success. Its video retrieval capabilities also streamline film analysis by providing direct access to specific plays and metadata. Below, a defensive coach uses *GridMind* to prepare for a game against the Ravens:

```
# USER: What is the offensive weakness of the Baltimore Ravens in the 2024 NFL regular season?
-----------------------------------------------------------------------------
# GRIDMIND: The Baltimore Ravens' offensive weakness in the 2024 NFL regular season, as of week 10, is their run blocking. They are ranked 19th out of 32 teams in run blocking, indicating that their performance in this area is below average compared to other teams in the league.
-----------------------------------------------------------------------------
# USER: What are the mismatches between the Minnesota Vikings' defense and Baltimore Ravens' offense in the 2024 NFL regular season?
-----------------------------------------------------------------------------
# GRIDMIND: The Minnesota Vikings' defense may struggle to contain the Baltimore Ravens' passing game, as the Ravens have the top passing composite score in the league, indicating a highly effective and efficient passing offense. This could pose a significant challenge for the Vikings' pass coverage, which ranks 17th in the league. Additionally, the Ravens have the 3rd ranked rushing composite score, highlighting their strong ground game, while the Vikings' rush defense ranks 28th, indicating a potential vulnerability against the run[2].
-----------------------------------------------------------------------------
```

---

[2] This response from *GridMind* is informed by several proprietary team evaluation metrics developed by Telemetry Sports.

### 6.4. Front Office Decision-Making

*GridMind* provides essential insights for front-office operations, supporting roster management, trade evaluations, and contract decisions. For instance, a general manager might ask, "Which free agents have excelled in man-to-man coverage over the past two seasons?" *GridMind* delivers a detailed report including player stats, footage, and performance comparisons, helping decision-makers align choices with team strategies and financial goals. Below, a general manager assesses the trade value of Anthony Richardson:

```
# USER: What is Anthony Richardson's trade value?
--------------------------------------------------------------------------------
# GRIDMIND: Anthony Richardson, a quarterback for the Indianapolis Colts, has a True Wins
Above Replacement (tWAR) of 0.10, ranking 39th out of 48 quarterbacks in the league. This
places his trade value in the lower tier among quarterbacks.
--------------------------------------------------------------------------------
# USER: What is his market cap?
--------------------------------------------------------------------------------
# GRIDMIND: Anthony Richardson's market cap for 2024 is $7,725,916.
--------------------------------------------------------------------------------
# USER: How much space will that free up for the colts if he leaves?
--------------------------------------------------------------------------------
# GRIDMIND: If Anthony Richardson leaves the Indianapolis Colts, the Colts will free up the
following amounts of cap space:

2023: $6,180,733
2024: $7,725,916
2025: $9,271,099
2026: $10,816,283

It appears that there is no information available for the amount of cap space that would be
freed up in 2027[3].
--------------------------------------------------------------------------------
```

Through these applications, *GridMind* demonstrates its versatility and effectiveness across fan engagement, broadcasting, coaching, and front-office decision-making, enhancing the accessibility of data-driven insights throughout the sports industry.

# 7. Performance Evaluation

Evaluating *GridMind's* performance involves benchmarking its accuracy, response time, scalability, and overall efficiency, and comparing these metrics to existing methods in sports analytics. These benchmarks assess *GridMind's* ability to deliver high-quality, contextually relevant responses within an interactive, multi-agent environment.

### 7.1. Benchmarking Criteria
1. **Data Retrieval Accuracy:** This metric assesses the precision of *GridMind's* responses, focusing on the accuracy of MongoDB queries and multimedia retrieval. By comparing system responses with expected outcomes across test queries, accuracy scores confirm

---

[3] True Wins Above Replacement (tWAR) is a comprehensive player evaluation metric developed by Telemetry Sports.

*GridMind's* reliability in retrieving contextually relevant sports data. A closed alpha with ten participants provided subjective measurements, with users rating each response. This analysis of accuracy and user perception informed improvements in data querying and model interpretability. *GridMind* demonstrated 58% accuracy based on binary good/bad feedback, with participants encouraged to mark responses as "bad" if the answer was incorrect, poorly worded, or lacked desired details. Although *GridMind* may have responded correctly to prompts technically, this stringent evaluation helped maximize future model tuning for optimal outcomes.

2. **Response Time:** In sports analytics, minimizing response time is critical for real-time decision-making. While *GridMind's* primary design focus was scalability, it has demonstrated strong support for rapid query processing. Internal benchmarks show that when the entire pipeline is executed sequentially without parallelization, the average response time is 17.5 seconds. However, *GridMind's* architecture allows for substantial parallelization, which can reduce response time significantly. Figure 2 illustrates the distribution of query response times, with outliers in the 20+ second range typically resulting from more complex queries. These extended times reflect a deliberate prioritization of response quality over speed, though further optimizations to enhance performance are planned for future iterations.

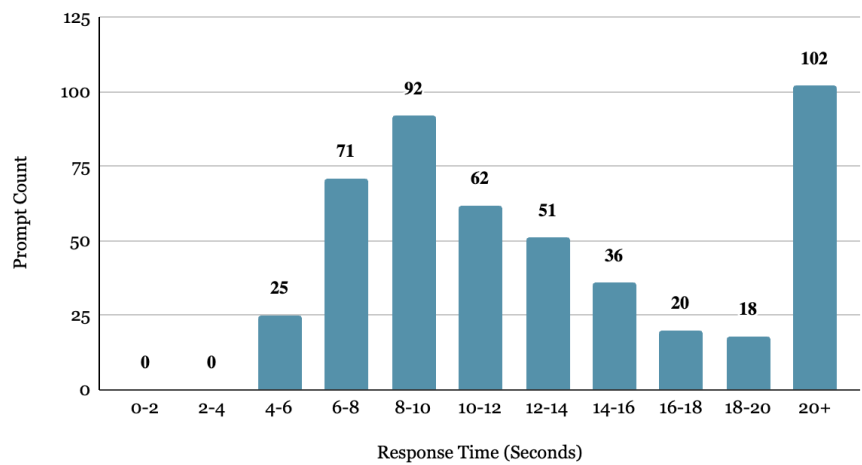

**Figure 2**: Distribution of query response time in seconds.

3. **Scalability:** *GridMind's* architecture has been tested for scalability under increasing data volumes and query complexity. Performance assessments from single-agent to multi-agent environments verify that *GridMind* maintains efficiency and response quality as data and user loads grow. Modular agents and message-passing enable robust scalability, essential for managing large sports datasets.

4. **Overall Efficiency:** *GridMind's* overall efficiency combines accuracy, response time, and scalability, reflecting its capacity to handle a broad range of sports-related queries quickly and accurately while optimizing computational resources. Efficiency is critical for real-time sports analytics, where large datasets and fast responses drive actionable insights. User perception of *GridMind* as a valuable assistant also influences engagement and perceived utility.

## 7.2. Comparative Analysis

Contextualizing *GridMind's* performance involves comparing its results with those from established sports analytics systems. *GridMind's* multimodal integration of video, text, and statistics sets it apart from traditional systems, which typically focus on single data types. This integration enables richer, more accurate insights in real-time, enhancing sports-related decision-making.

*GridMind's* approach aligns with the ways fans and analysts increasingly consume sports data, from quick updates on X (formerly Twitter) to aggregated news on Google and specialized insights from LLMs like ChatGPT, Gemini, and Claude. By integrating data from multiple sources, *GridMind* reflects these trends while enhancing accessibility to comprehensive insights across platforms.

# 8. Limitations

While *GridMind* showcases effective multi-modal data integration and real-time querying, several limitations challenge its performance in complex scenarios. These challenges stem primarily from the diversity of data formats, the complexity of query processing, the subjectivity in nuanced conversations, and the computational demands of large-scale, real-time analysis.

## 8.1. Handling Nuanced Queries

*GridMind* can struggle with highly nuanced or ambiguous queries, particularly when dealing with unstructured data from various sources or lengthy, multi-topic conversations. Differences in terminology across media—such as language used in commentary versus structured data—can impact retrieval accuracy and coherence. Enhanced NLP techniques, including domain-specific fine-tuning or training on sports lexicons, may improve *GridMind's* interpretation of language variations and context.

Certain topics also introduce subjectivity that affects user perception of accuracy. For example, in a closed alpha test, a participant provided negative feedback when *GridMind* returned a player position unfamiliar to them:

```
# USER: Build me the perfect team from the 2022 season.
-----------------------------------------------------------------------------
# GRIDMIND:
QB | Patrick Mahomes | KC | 5.79 tWAR |
RB | Josh Jacobs | LV | 0.33 tWAR |
WR | Justin Jefferson | MIN | 0.43 tWAR |
TE | Travis Kelce | KC | 0.23 tWAR |
OT | Trent Williams | SF | 0.17 tWAR |
IOL | Creed Humphrey | KC | 0.25 tWAR |
EDGE | Myles Garrett | CLE | 0.61 tWAR|
DT | Quinnen Williams | NYJ | 0.47 tWAR |
ILB | Fred Warner | SF | 0.21 tWAR |
OB-LB | Jaquan Brisker | CHI | 0.11 tWAR |
...
-----------------------------------------------------------------------------
# USER: (rated negatively via thumbs down) OB isn't a position in Football - Would assume
this should be DB/LB instead.
-----------------------------------------------------------------------------
```

This feedback highlights how *GridMind's* use of granular positional labels, like Off-Ball Linebackers (OB-LB), may be unfamiliar to users, impacting perceived accuracy. Future development should account for such subjective aspects to refine *GridMind's* responses in scenarios inviting interpretation.

### 8.2. Multi-Modal Integration Constraints
*GridMind's* ability to integrate multi-modal data depends heavily on the quality and consistency of metadata for each data type. Incomplete or inconsistent metadata—such as missing tags in audio files or low-resolution audio—can limit *GridMind's* effectiveness in making cross-modal connections. Enhancing data preprocessing, with additional metadata enrichment or automated tagging, could improve *GridMind's* accuracy and cohesion across formats.

### 8.3. Scalability and Computational Overhead
Scaling *GridMind* to continuously process extensive NFL data while maintaining real-time performance is computationally demanding. Multi-modal embeddings and dense retrieval methods require significant memory and processing power, especially for large datasets and complex queries. Techniques such as model pruning, caching, and GPU-accelerated similarity searches could enhance scalability without sacrificing performance.

### 8.4. Dependency on Training Data Quality
*GridMind's* parsing and data integration accuracy depend closely on the quality and representativeness of its training data. Low-quality or biased data can reduce effectiveness with complex or diverse queries. Our rollout strategy includes staged releases, following the closed alpha model, to introduce unbiased users with fresh perspectives, providing valuable feedback for training. Future efforts should focus on data curation to ensure *GridMind* interprets a wide range of sports-related queries accurately.

### 8.5. Model Training and Tuning for High-Volume Query Handling
As the volume of prompts grows, manual evaluation becomes inefficient. To streamline tuning, we employ a hybrid approach that combines manual evaluation with automated scoring.

1. **Manual and Automated Evaluation:** Initially, each prompt undergoes manual evaluation to refine the LLM, but this alone is unsustainable as prompt volumes increase. An automated scoring system now assesses prompt complexity and prioritizes evaluation accordingly.

2. **Challenge Scoring:** The system assigns a challenge score to each prompt based on its complexity. This score reflects the query structure; more complex graphs indicate higher difficulty and potential error, enabling *GridMind* to prioritize complex, multi-domain queries.

3. **Hard Example Mining:** This challenge-based scoring approach emphasizes difficult prompts, ensuring high accuracy on complex queries. This prioritization optimizes the tuning process, allowing *GridMind* to adapt to increasingly intricate question patterns over time.

# 9. Future Work

*GridMind's* modular, multi-modal architecture lays a strong foundation for further advancements, enabling enhancements that could optimize scalability, deepen data integration, and broaden applications beyond NFL analytics. Key areas for future development will enhance *GridMind's* performance and adaptability in increasingly data-rich environments.

## 9.1. Enhanced Scalability and Modularity

To meet growing demand for real-time sports analytics, scalability will be vital for *GridMind's* ability to handle higher query volumes and complex data requirements. Future updates could incorporate parallel processing, model pruning, and adaptive load-balancing to sustain rapid response times under heavy data loads. Introducing features like typing emulation, where *GridMind* displays responses as they are processed, could serve dual purposes: providing additional processing time while enhancing user experience with a more human-like interaction. Further modular extensions, such as adding new agents to manage diverse data formats, could expand *GridMind's* ability to handle a wider range of queries simultaneously.

## 9.2. Advanced Query Interpretation

Refining *GridMind's* natural language understanding could improve its accuracy in handling nuanced queries. Future work may leverage enhanced context injection and domain-specific adaptation to deepen query interpretation. Incorporating advanced NLP models fine-tuned on sports commentary or adopting sports-specific language datasets could increase precision with complex queries. Enhancements in context injection, including relevant game scenarios, player roles, or recent performance metrics, could further enrich responses. Offering distinct response styles (e.g., "coach," "fan," or "broadcaster") would allow users to select perspectives, while also providing insights for developers to tailor performance to diverse expectations. Adopting techniques like transfer learning from sports-specific lexicons would strengthen *GridMind's* understanding of context-sensitive terms, resulting in more nuanced interpretations of user queries.

## 9.3. Expanding Data Sources and Modalities

Integrating additional data sources, such as physiological metrics, social media sentiment, or real-time telemetry, could further enhance *GridMind's* analytical depth. Expanding to process unstructured data—such as real-time images, audio, or social media feeds—would support a more comprehensive range of insights. Although unaligned data sources add complexity, multi-modal transformers, as noted by Tsai, Y.-H. H., Bai, S., et al. [29], could allow *GridMind* to synthesize data across modalities more effectively, especially when managing asynchronous inputs at scale.

## 9.4. Broader Applications Across Sports and Domains

While currently tailored for NFL analytics, *GridMind's* architecture can be adapted to other sports and domains where complex data integration is valuable. Expanding to new sports, like basketball, soccer, or esports, would require sport-specific training data but could open new opportunities for real-time, multi-modal analysis across athletic contexts. Additionally, *GridMind* could extend to fields beyond sports—such as finance, healthcare, or logistics—demonstrating its versatility as a data-driven decision support system for various data-intensive industries.

# 10. Conclusion

This paper presents *GridMind*, a multi-agent framework designed to unify diverse NFL data sources, addressing the challenges of integrating structured statistics, textual insights, and multimedia content. By combining a modular, multi-agent architecture with retrieval-augmented generation, *GridMind* delivers complex, context-rich insights through an intuitive interface. Rigorous benchmarking underscores its utility across fan engagement, coaching, broadcasting, and front-office decision-making, setting a new standard for multi-modal sports analytics.

*GridMind's* architecture, with specialized agents, enables efficient query parsing, data retrieval, and response synthesis, offering a scalable solution tailored to real-time sports analytics. Benchmarking results demonstrate its effectiveness in delivering precise, comprehensive responses across a range of applications, from enhancing fan engagement and broadcasting to informing coaching strategies and front-office decisions.

While *GridMind* shows strong performance, areas for future research include advancing query interpretation, integrating broader data sources, and optimizing scalability to manage larger datasets. These enhancements will expand *GridMind's* applicability within sports analytics and other data-rich domains, reinforcing its role as a versatile tool for sophisticated multi-modal data integration.

In summary, *GridMind* represents a significant advancement in sports data analysis, seamlessly unifying diverse data sources to provide actionable insights for fans, analysts, and professionals alike. By making data-driven insights more accessible, *GridMind* promotes a deeper understanding of the game and supports informed decision-making across the NFL and beyond.